# Reputation-based Incentive Protocols in Crowdsourcing Applications

Yu Zhang
Department of Electrical Engineering
University of California, Los Angeles
Los Angeles, CA 90095, USA
yuzhang@ucla.edu

Mihaela van der Schaar
Department of Electrical Engineering
University of California, Los Angeles
Los Angeles, CA 90095, USA
mihaela@ee.ucla.edu

*Abstract*—Crowdsourcing websites (e.g. Yahoo! Answers, Amazon Mechanical Turk, and etc.) emerged in recent years that allow requesters from all around the world to post tasks and seek help from an equally global pool of workers. However, intrinsic incentive problems reside in crowdsourcing applications as workers and requester are selfish and aim to strategically maximize their own benefit. In this paper, we propose to provide incentives for workers to exert effort using a novel game-theoretic model based on repeated games. As there is always a gap in the social welfare between the non-cooperative equilibria emerging when workers pursue their self-interests and the desirable Pareto efficient outcome, we propose a novel class of incentive protocols based on social norms which integrates reputation mechanisms into the existing pricing schemes currently implemented on crowdsourcing websites, in order to improve the performance of the non-cooperative equilibria emerging in such applications. We first formulate the exchanges on a crowdsourcing website as a two-sided market where requesters and workers are matched and play gift-giving games repeatedly. Subsequently, we study the protocol designer's problem of finding an optimal and sustainable (equilibrium) protocol which achieves the highest social welfare for that website. We prove that the proposed incentives protocol can make the website operate close to Pareto efficiency. Moreover, we also examine an alternative scenario, where the protocol designer aims at maximizing the revenue of the website and evaluate the performance of the optimal protocol.

## I. Introduction

Soliciting solutions to various tasks using online labor markets has become increasingly popular in recent years. The term "crowdsourcing", a form of "peer production" that outsources works to a large group of people [1], was recently coined to refer to such approaches. Numerous crowdsourcing websites, such as Yelp [2], Yahoo! Answers [3] and Amazon Mechanical Turk [4], can be viewed as systems where small tasks (typically on the order of minutes or seconds) and performed in exchange for rewards awarded to the users who performed them. Examples of tasks found on crowdsourcing websites are the graphical design of logos, the creation of a marketing plan, the answering of a computer programming question, etc. For illustration purposes, we differentiate users on a crowdsourcing website by two different types. A user who posts tasks is called a *"requester"*, and a user who works on a posted task is called a *"worker"*. Many crowdsourcing websites exhibit similar structures – a task is described and posted by a requester together with the associated reward. Workers submit solutions to the task, and the requester selects a subset of submissions (usually the first one that solves the task) with the corresponding workers granted the reward [5].

Despite the success and the perceived promise of online labor markets, crowdsourcing websites face a serious practical challenge: providing appropriate incentives for workers to participate and well-perform in tasks. More concretely, a requester needs to establish sufficient rewards to attract workers' contributions when workers do not solve tasks solely for altruistic motivations [5][6][10][11]. Such studies demonstrate that designing incentive mechanisms to encourage workers' contributions is crucial to maintain the performance of crowdsourcing websites.

Many of the incentive mechanisms on crowdsourcing websites rely on monetary rewards in the form of micropayments [5][7][8][9]. The requester pays workers in the form of cash upon the completion of a task. However, the monetary incentive mechanism possesses several disadvantages which prevent it from being easily deployed on crowdsourcing websites. The first problem associated with the payment system is the absence of associated effective pricing schemes. Due to the heterogeneity in tasks and requesters, the existing pricing scheme often provides competitive advantages to the requesters with high budgets [10]. Secondly, such pricing schemes often deploy auctions to set the price [7], which may result in high delay and implementation complexity in order to achieve desirable resource allocations and often cause "currency inflation" on the crowdsourcing websites. Moreover, it is also worth noting that to deploy such pricing schemes usually necessitates the usage of complicated and reliable financial accounting, which is difficult to design and deploy in large-scale online websites. Finally and most importantly, the current pricing schemes cannot solve the social dilemma existing between workers and requesters: if the payment of a task is ex-ante, which means that the requester pays before the task starts, a worker always has the incentive to take the payment and provides no effort to solve the task, a behavior commonly known as "Free-riding" [13]; whereas if the payment is ex-post, which means that the requester pays after the task is solved, he always has the incentive to refuse the payment to workers by lying about the outcome of this task, a behavior commonly known as "false-reporting" [13].

In this paper we address these challenges by designing a novel class of incentive protocols which rely on reputation mechanisms and only require a simple flat-rate pricing scheme. We assume that the payment is performed ex-ante on a crowdsourcing website, and focus mainly on the "free-riding" of workers. However, our protocol and system design can be straightforwardly extended to address the "false-reporting" problem of requesters.

We assign each worker on a crowdsourcing website with a reputation which is gained based on its past behavior on the website. A differential reputation-based reward and punishment scheme is performed by some intermediary (e.g. the website administrator): a worker with a higher reputation will be given a higher chance to participate in tasks and receives more payments upon the completion of a task. Since a worker with a high reputation is treated preferentially, incentives to contribute are provided to workers in order to build up high reputations.

Different from the traditional models in the design of pricing schemes where the interactions between workers and requesters are usually formalized as a one-shot game [7][8], we capture the repeated interactions between requesters and workers within the formalism of repeated games. To formalize the reputation scheme, *social norms* [15], which consist of a social strategy and a reputation scheme, are introduced to regulate the behavior of workers. By designing an appropriate social norm to punish workers' deviations from the selected social strategy, we rigorously analyze how workers' equilibrium behaviors will be influenced by the benefit and cost of tasks as well as their patience and show the tradeoff between the efficiency on the social welfare and the workers' incentives. Subsequently, we quantify the sufficient and necessary conditions under which workers will find in their own self-interest to comply with the prescribed social strategy. We then define and solve the protocol design problem of selecting an optimal social norm that maximizes the social welfare of the website. Using this analysis, we also discuss how the protocol design should be adjusted in order to maximize the revenue of the website. In the numerical simulations, we also consider the requesters' strategic behavior and investigate how the requesters' incentives will influence the optimal protocol design and the resulting problem of designing the optimal pricing schemes.

In summary, the novelty of our work lies in the following aspects:

(1) We explicitly formalize the interactions between workers and requesters using a rigorous repeated game framework to analyze the time-dependency of their current and future behavior.

(2) We propose a novel class of incentive protocols based on social norms, which is simple to design and flexible in implementation, in order to incentivize workers to participate and contribute. We prove that our protocol can prevent the "free-riding" problem of workers and incentivize them to contribute their efforts. This paper, according to our knowledge, is the first work that integrates payment and reputation mechanisms in a rigorous manner to design protocols for crowdsourcing applications.

(3) Different from most ad-hoc protocol designs for crowdsourcing applications that are empirical and rely on trial-and-error approaches, we rigorously analyze the structure of equilbria, which is critical in understanding the relationships between our designed protocol, the intrinsic market characteristics (e.g. rewards, costs, workers' patience), and the workers' incentives on a crowdsourcing website.

(4) Using simulations, we numerically analyze the incentive problem of false-reporting on requesters, which is also neglected in current protocol designs.

The remainder of the paper is organized as follows. In Section II, a rigorous analytical framework is proposed to analyze the crowdsourcing website. In Section III, we analyze the non-cooperative equilibria of this model and discuss how our protocol design influences workers' incentives. Section IV investigates the revenue maximization problem. After showing the simulation results in Section V, we conclude the paper in Section VI.

## II. SYSTEM MODEL

### A. Setup

A crowdsourcing website usually covers a wide variety of tasks. For example, Yahoo! Answers has 25 categories ranging from "Computers & Internet" to "Travel" to "Family & Relationships" to "Health" [3]. In this work, we explicitly analyze the workers' incentives in a particular type of tasks, where questions are homogeneous in terms of the expertise and efforts needed to solve them. It should be noted that our analysis does not lose any generality by imposing this assumption. Since the incentive protocols designed in this paper can be easily extended to the scenario with multiple types of tasks by designing reputation systems and mechanisms for each individual type, respectively.

To model the large user population, we utilize the widely-used continuum model. Meanwhile, we assume that there is a larger population of requesters than that of workers. For simplicity, we assume that workers are long-lived on the website.

When a requester generates a task, it is posted on the website with the price paid for completing it to be specified. We assume that the pricing scheme is exogenously determined (e.g. by the website owner) [10]. Regarding the fact that we consider tasks to be homogeneous, a flat-rate pricing scheme is adopted here, i.e. a requester pays a fixed price $p$ for each task. Workers are randomly selected for the different posted tasks. The selection is uniformly random, such that all posted tasks have an equal probability to be chosen by a worker. However, the selection is only performed among the workers who have the necessary knowledge and expertise to solve the particular type of task. This assumption is not restrictive since our analysis shows that the designed protocol is able to automatically block workers who are not capable to solve this type of tasks out of the system. Hence, all workers are equally capable in solving tasks posted on this website. The website is modeled as a discrete-time system where time is divided into periods. By selecting the length of a period as the normal amount of time for a worker to solve a task, each worker can perform only one task in one period. The probability that more than one worker are devoting to the same task is small and assumed to be 0 in our analysis.

### B. Stage Game

The interaction between a worker and a requester in a task, which is defined as a "*transaction*", can be modeled as an asymmetric gift-giving game [14]. We assume that the payment of the requester is ex-ante, i.e. he has to submit the payment once the worker starts to working on this task [1]. In Section V, we explicitly discuss the incentive of requesters when they can strategically choose to submit the payment or not in a transaction. The payment is shared by the worker and the website. Particularly, the worker receives $\lambda p$ where $\lambda < 1$ as its reward, and the website charges an amount $(1-\lambda)p$, which can be regarded as the maintenance cost or the usage fee of the website. The payment sharing ratio $\lambda$ is explicitly determined by the protocol designer.

Because the worker receives the payment in advance, he can strategically choose his action, i.e. determine the level of effort devoted to this task. The worker's action will not only impact his own utility, but also that of the requester. For simplicity, we assume that a worker's action $a$ is chosen from a binary set $\mathcal{A} = \{H, L\}$ [2], where $H$ stands for "High level of effort", whereas $L$ stands for "Low level of effort". The utility matrix of one transaction is illustrated in Table 1, which is specified as follows.

---
[1] This model is utilized in several crowdsourcing websites [1], where the payment of the requester is click-based, i.e. it has to pay for each click that workers submit to its post.

Table 1. The utility matrix of one transaction

|  | Worker | |
|---|---|---|
|  | H | L |
| Requester | $V - p$, $\lambda p - c$ | $-p$, $\lambda p$ |

If $a = H$, the worker consumes a cost $c$ for solving this task. The task is then solved and the requester receives a benefit of $V$. As we assume that tasks are homogeneous and workers are identical in terms of expertise and knowledge, we consider that $c$ and $V$ are constant for each task in our formal analysis. However, our proposed framework can be extended to take task-dependent costs and benefits into consideration. We assume that $V > c$ such that the task-solving process is socially valuable. Upon its completion, the task is removed from the website.

If $a = L$, the worker free-rides by taking the payment and consuming a low cost, which is approximated by 0 here, and the requester receives no benefit [3]. The task is not solved and remains open for future workers.

The social welfare $U$ is proportional to the sum utility of all users (requesters and workers) in one period and it is optimized when all workers devote a high level of efforts into their tasks. Nevertheless, the dominant strategy for a worker is to play $a = L$ in order to maximize its utility myopically, which gives rise to an undesirable social welfare.

*C. Social norms*

To formalize the repeated interactions among workers and requesters, a general repeated game formulation is deployed. An incentive protocol based on the idea of social norm is designed to improve the inefficiency of the myopic equilibrium. The formal definition of a social norm can be found in [15]. In this paper, it defines the rules that the manager of the website, which is called "website administrator" here, uses to reward or punish workers in order to regulate their behavior. It should be noted that a website administrator only takes charge of implementing the protocol designed by the protocol designer. Whether these two roles are played by the same person on a website does not affect our analysis. We differentiate them only for the purpose of illustrations.

Formally, a social norm $\kappa$, which is design by the protocol designer, is composed of a social strategy $\sigma$, a reputation scheme $\tau$, and a reputation set $\Theta$.

In the repeated game, each worker is tagged with a reputation $\theta$ representing its social status. $\theta$ is a natural number from the finite set $\Theta = \{0, 1, 2, \cdots, K_\kappa\}$, where $K_\kappa$ represents the size of this set. A high reputation relates to a worker's good social status, which reflects his good behavior on solving tasks in the past. The reputation of each worker is maintained by the administrator. It is updated depending on the report of the requester about the outcome of the transaction. The detailed scheme and process of reputation updating will be introduced later.

$\sigma$ is a *reputation-based strategy*, which is represented by a mapping: $\sigma : \Theta \to \mathcal{A}$, where $\Theta$ represents the worker's reputation, and $\mathcal{A}$ represents the approved action of the worker at a particular reputation.

$\tau$ serves as *the reward and punishment system* in the social norm, and it specifies how a worker's reputation should be updated based on its actions in the transactions that it is engaged. In our framework, $\tau$ updates a worker's reputation at the end of each transaction. Specifically, $\tau$ is represented by a mapping $\tau : \Theta \times \mathcal{A} \to \Theta$, where $\mathcal{A}$ is the worker's reported action in a transaction. If the requester thinks that the task is solved, the worker is rewarded with his reputation increased; on the other hand, if the task is not solved, the worker is punished and his reputation decreases. Since a requester cannot benefit by falsely reporting the outcome as the payment is ex-ante, he is always truth-telling. However, there would be some discrepancy between the worker's devoted effort level and the requester's perception on the outcome. Specifically, the requester might be unsatisfied with the outcome and report the task as unsolved even if the worker has devoted a high level of effort; whereas the requester might regard the task as being solved when the worker devoted few effort and played $a = L$. Such scenarios are assumed to happen in a transaction with a small probability $\alpha \ll 1$.

To encourage the workers to contribute and mitigate free-riding, we restrict our attention to a simple class of threshold-based social strategies $\Gamma$. A worker will be isolated by the administrator and is forbidden to interact with requesters and participate in any task if his reputation is low. In this case, the social norm does not require the worker to do anything. On the contrary, the administrator "activates" the worker by allowing him to participate in tasks if his reputation is high, and the social norm require the worker to devote a high level of efforts in his transactions. Every social strategy $\sigma \in \Gamma$ can be characterized by a service thresholds $h(\sigma) \in \{1, \cdots, K_\kappa\}$ [4] and is specified as follows:

$$\sigma(\theta) = \begin{cases} H & if \ \theta \geq h(\sigma) \\ L & if \ \theta < h(\sigma) \end{cases}. \quad (1)$$

Let $h_\kappa$ denote the threshold of the selected social strategy and is called as the social threshold, the reputation scheme we adopted here is as follows:

$$\tau(\theta, a) = \begin{cases} \min\{K_\kappa, \theta + 1\} & if \ a = H \ and \ \theta \geq h_\kappa \\ \theta - 1 & if \ a = L \ and \ \theta \geq h_\kappa + 1 \\ 0 & if \ a = L \ and \ \theta = h_\kappa \\ \theta + 1 & if \ \theta < h_\kappa \end{cases}. \quad (2)$$

Under this reputation scheme, when a worker is active (i.e $\theta \geq h_\kappa$), his reputation increases by 1 while not exceeding $K_\kappa$ after one period, if his requester reports the task as being solved. Once the task is being reported as unsolved, the worker's reputation drops by 1 as a warning from the administrator. When the worker's reputation falls to $h_\kappa$ and he receives another negative feedback from his requester, he is isolated with his reputation falling to 0. During the window of isolation, the worker's reputation goes back by one point in each period until reaching $h_\kappa$ when he is activated again. Hence, this reputation scheme brings up the punishment to workers with an isolation of $h_\kappa$ periods. For illustration purposes, we call a worker who complies with the social norm as a "compliant worker" and a worker who deviates from the social norm as a "non-compliant worker".

---

[2] Our framework can also be applied to the case when $a$ has multiple levels without changing the analysis.

[3] This models the scenario when the requester receives a low benefit and the worker submits a low effort. We take the extreme value of 0 only for the simplicity for the analysis.

[4] It should be noted here when the threshold is lower than 1 or higher than $K_\kappa + 1$, all workers are treated equally and the free-riding problem cannot be solved in this case.

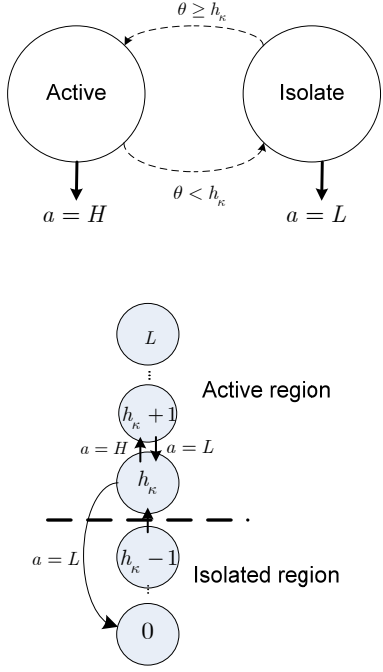

Figure 1  The schematic representation of a social norm

We make several remarks here:

(1) The design purpose of the social norm is to enforce a worker to play with the social strategy $\sigma$, in which $a = H$ is played whenever he is active.

(2) If there is a worker who does not have the necessary expertise to solve the tasks. He will be isolated for most of the time and will not affect the social welfare.

(3) If there are multiple sections or discussion boards on a website focusing on different types of tasks. The above incentive scheme can be simply extended by designing one separate set of reputation and a related reputation scheme for each type, respectively. In this case, a worker maintains a vector of reputations, where each element reflects his social status in the section of a particular type.

A schematic representation of a social norm is provided in Figure 1 with the upper one illustrating the decision process of a social strategy, and the lower one illustrating the decision process of a reputation scheme.

It is worth noting that an isolated worker's reputation transition across periods is not affected by the error probability $\alpha$ since he is not engaged in any transaction. On the contrary, an active worker's reputation has a positive probability of not being corrected updated. A worker's reputation transition probability across periods will be explicitly calculated in the next section.

*D. Utilities*

In this section, we discuss the utility of compliant workers. An isolated worker receives a utility of 0 in one period and an active worker receives $\lambda p - c$, i.e.

$$v_\kappa(\theta) = \begin{cases} \lambda p - c, & if \ \theta \geq h_\kappa \\ 0, & if \ \theta < h_\kappa \end{cases}. \quad (3)$$

Obviously, we need $\lambda \geq \dfrac{c}{p}$ to provide workers with correct incentives, which is assumed throughout the paper. We use the infinite-horizon discounted sum criterion to evaluate a worker's expected long-term utility as the sum of his one-period utility in the current period and his discounted expected future utility starting from the next period. Let $p_\kappa(\theta' \mid \theta)$ denote the transition probability of a compliant worker's reputation across periods under the social norm $\kappa$, it can be determined as follows

$$p_\kappa(\theta' \mid \theta) = \begin{cases} 1-\alpha, & \theta \geq h_\kappa \ and \ \theta' = \min\{K_\kappa, \theta+1\} \\ \alpha, & \theta \geq h_\kappa \ and \ \theta' = \theta - 1 \\ \alpha, & \theta = h_\kappa \ and \ \theta' = 0 \\ 1, & \theta < h_\kappa \ and \ \theta' = \theta + 1 \\ 0, & otherwise \end{cases} \quad (4)$$

Therefore, a compliant worker's expected long-term utility in the repeated game starting from any period $t_0$, can be expressed as

$$\begin{aligned} v_\kappa^\infty(\theta^{(t_0)}) &= \mathbb{E}\left[\sum_{t=t_0}^\infty \delta^t v_\kappa(\theta^{(t)})\right] \\ &= v_\kappa(\theta^{(t_0)}) + \delta \sum_{\theta'} p_\kappa(\theta' \mid \theta^{(t_0)}) v_\kappa^\infty(\theta') \end{aligned} \quad (5)$$

where $\delta \in [0,1)$ is the discount factor describing a worker's patience, which represents the weight that a worker gives to his utility that can be received in the future. Since all workers are long-lived on the website, we assume that they all have the same discount factor.

The social welfare of the website depends on the reputation distribution of the workers' population. The reputation distribution in one period is denoted by $\boldsymbol{\eta} = \{\eta(\theta)\}_{\theta=0}^L$, with each term $\eta(\theta)$ representing the fraction of workers in the workers' population holding a reputation $\theta$. Due to the reputation update (performed by the administrator at the end of each period), $\boldsymbol{\eta}$ evolves dynamically over time. Since we are interested in the long-term utilities of workers, we study the stationary distribution $\boldsymbol{\eta}_\kappa$ in the long run when all workers complying with the social norm, which is defined as follows:

$$\begin{aligned} \eta_\kappa(K) &= (1-\alpha)\eta_\kappa(K_\kappa) + (1-\alpha)\eta_\kappa(K_\kappa - 1) \\ \eta_\kappa(\theta) &= (1-\alpha)\eta_\kappa(\theta-1) + \alpha\eta_\kappa(\theta+1), \ h_\kappa + 1 \leq \theta \leq K_\kappa - 1 \\ \eta_\kappa(h_\kappa) &= \eta_\kappa(h_\kappa - 1) + \alpha\eta_\kappa(\theta+1) \\ \eta_\kappa(\theta) &= \eta_\kappa(\theta-1), \ 1 \leq \theta \leq h_\kappa - 1 \\ \eta_\kappa(0) &= \alpha\eta_\kappa(h_\kappa) \end{aligned}$$

(6)

Therefore, the social welfare of the website is defined as the expected one-period utility averaged over all workers and all

requesters who are engaged in transactions when the reputation distribution is stationary [5]

$$U_\kappa = \sum_{\theta < h_\kappa} \eta_\kappa(\theta) v_\kappa(\theta) + \sum_{\theta \geq h_\kappa} \eta_\kappa(\theta)(v_\kappa(\theta) + V - p). \quad (7)$$

III. OPTIMAL DESIGN OF SOCIAL NORM BASED PROTOCOLS

*A. Defining sustainable protocols*

Workers are rational and self-interested, they will comply with the prescribed social norm if and only if it is in their self-interests to do so, i.e. they cannot benefit in terms of their long-term utility upon deviations. Hence, the incentive of active workers has to be investigated when designing the protocol, in order to determine when compliance is individually optimal to all of them [6].

Because a worker's long-term utility possesses a recursive structure according to (5), individual optimality can be determined by the one-shot deviation principle [12]: if a worker cannot benefit upon a single deviation from the social strategy, i.e. he deviates in one period but complies with the social strategy subsequently, he also cannot gain by deviating to any strategy other than $\sigma$. On the basis of the one-shot deviation principle, we define the social norm equilibrium as follows:

**Definition 1 {Social Norm Equilibrium}**. A social norm $\kappa = (\sigma, \tau)$ is in equilibrium if and only if

$$c \leq \delta \left( \sum_{\theta'} p_\kappa(\theta' | \theta) v_\kappa^\infty(\theta'') - \sum_{\theta'} p(\theta' | \theta, a) v_\kappa^\infty(\theta') \right). \quad (8)$$

where $p(\theta' | \theta, a)$ represents a worker's reputation transition probability under $\kappa$ when he chooses an action $a$. Clearly, we have $p_\kappa(\theta' | \theta) = p(\theta' | \theta, \sigma(\theta))$ for all $\theta$ and $\theta'$.

A protocol consists a social norm $\kappa$ and a payment sharing ratio $\lambda$, the sustainability of a protocol is correspondingly defined as follows.

**Definition 2 {Sustainable Protocol}**. A protocol $(\kappa, \lambda)$ is sustainable if and only if $\kappa$ is a social norm equilibrium.

The equilibrium condition (8) is analyzed in details below.

For a compliant worker of reputation $\theta \geq h_\kappa + 1$, he plays $a = H$ and consumes a cost $c$. His reputation increases to $\min\{K, \theta + 1\}$ with the probability $1 - \alpha$ and decreases to $\theta - 1$ the probability $\alpha$. Hence his long-term utility is
$V_\theta(H) = \lambda p - c + \delta \left[ (1 - \alpha) v_\kappa^\infty(\min\{K, \theta + 1\}) + \alpha v_\kappa^\infty(\theta - 1) \right]$.
On the contrary, if the worker submits a one-period deviation and plays $a = L$, he saves the immediate cost $c$, while the probabilities of his reputation to increase and decrease become $\alpha$ and $1 - \alpha$, respectively. The long-term utility then becomes
$V_\theta(L) = \lambda p + \delta \left[ (1 - \alpha) v_\kappa^\infty(\theta - 1) + \alpha v_\kappa^\infty(\min\{K, \theta + 1\}) \right]$.
According to (8), the worker has no incentive to deviate if and only if $V_\theta(H) \geq V_\theta(L)$.

---
[5] Since requesters whose tasks are not being solved can be regarded as being engaged in transactions with isolated workers with a zero utility in this period, our formulation does not lose any generality.

[6] It should be noted here that the incentive for isolated workers do not need to be considered since they do not participate in any task and will comply with the social norm by default.

By applying similar analysis to workers of reputation $h_\kappa$, we derive the following condition for a social norm to be in equilibrium:

$$0 \leq \delta(1 - 2\alpha) \left[ v_\kappa^\infty(\min\{K_\kappa, \theta + 1\}) - v_\kappa^\infty(\theta - 1) \right] - c, \text{ if } \theta \geq h_\kappa + 1,$$
$$0 \leq \delta(1 - 2\alpha) \left[ v_\kappa^\infty(\min\{K_\kappa, h_\kappa + 1\}) - v_\kappa^\infty(0) \right] - c, \text{ if } \theta = h_\kappa.$$
$$(9)$$

In our subsequent analysis, we refer to the RHS of (9) as workers' incentives. A positive incentive guarantees a worker's compliance to the social norm, and a protocol is sustainable if and only if workers have positive incentives at any $\theta \geq h_\kappa$.

*B. Problem formulation*

In this section, we discuss the design of a sustainable protocol that maximizes the social welfare of the website. Based on the above analysis, the protocol design can be simplified to the selection on three parameters: the size $K_\kappa$ of the reputation set, the social threshold $h_\kappa$, and the payment sharing ratio $\lambda$. This is formalized in (10) (we call this problem "social welfare optimization"). Nevertheless, it is worth noting that the optimal social welfare also depends on the parameters $c$, $p$, $\delta$, and $\alpha$, which are assumed to be fixed and cannot be designed by the protocol designer. In the remainder of this paper, these parameters are referred as intrinsic parameters of the website.

$$\max_{(K_\kappa, h_\kappa, \lambda)} U_\kappa = \sum_{\theta < h_\kappa} \eta_\kappa(\theta) v_\kappa(\theta) + \sum_{\theta \geq h_\kappa} \eta_\kappa(\theta)(v_\kappa(\theta) + V - p)$$

subject to

$$c \leq \delta(1 - 2\alpha) \left[ v_\kappa^\infty(\min\{L, \theta + 1\}) - v_\kappa^\infty(\theta - 1) \right], \text{ if } \theta \geq h_\kappa + 1,$$
$$c \leq \delta(1 - 2\alpha) \left[ v_\kappa^\infty(\min\{L, \theta + 1\}) - v_\kappa^\infty(0) \right], \text{ if } \theta = h_\kappa.$$
$$(10)$$

In (10), we explicitly focus on sustainable protocols without considering all possible protocols in the design space. To justify such restriction, we first prove in the following theorem that the social welfare that can be achieved by a social norm that is not in equilibrium is always smaller than the optimal value of (10) and hence, it is only necessary to consider sustainable protocols in order to maximize the social welfare. The detailed proof is deferred to [16].

**Theorem 1.** Any social norm that does not satisfy (9) delivers a social welfare that is smaller than the optimal value of (10).

*Proof*: See [16, Appendix A]. ∎

*C. The design of optimal sustainable protocols*

The social welfare optimization requires the joint consideration on how a protocol impacts the social welfare and the workers' incentives of compliance. In this section, we separately analyse the impacts of three design parameters $(K_\kappa, h_\kappa, \lambda)$ in order to characterize the optimal design, which is denoted as $(K^*, h^*, \lambda^*)$.

We first analyse the payment sharing ratio $\lambda$. Since an increasing $\lambda$ only reduces the revenue of the website without hurting workers' utilities, it is easy to see that given any social norm $\kappa$, the social welfare and the workers' incentives are both maximized when $\lambda = 1$.

**Proposition 1.** Given a social norm $\kappa$, $\lambda^* = 1$ is always the optimal solution of (10).

*Proof*: See [16, Appendix A]. ∎

In the remainder of our design, we set $\lambda = 1$ by default without further notice.

Next, we focus on the design of the social norm. We first analyze the relationship between the social welfare $U_\kappa$ and $(K_\kappa, h_\kappa)$ by assuming workers' compliance, i.e. the incentive constraints (9) are not explicitly considered.

Given $h_\kappa$, the fraction of isolated workers in the workers' population decreases with an increasing $K_\kappa$. Hence, $U_\kappa$ will increase with more workers' participations. On the other hand, given $K_\kappa$, a larger $h_\kappa$ implies a larger window of isolation and thus more severe punishment on workers, which reduces the workers' participation ratio in tasks and thus $U_\kappa$. Therefore, without incentive constraints (9), it is always optimal to select $h_\kappa = 1$ and the largest $K_\kappa$ that is allowed by the website in order to maximize the social welfare. The design problem now is transformed into the selection for the smallest $h_\kappa$ and the largest $K_\kappa$ with which the incentive constraints (9) are satisfied. These observations are summarized in the following proposition.

**Proposition 2.** $U_\kappa$ monotonically increases with $K_\kappa$ and monotonically decreases with $h_\kappa$.

*Proof*: See [16, Appendix A]. ∎

The analysis on the relationship between the workers' incentives and $(K_\kappa, h_\kappa)$ is more complicated, since we need to consider the time dependency between workers' current and future utilities. To facilitate the analysis, we define the marginal utilities of workers as

$$\Delta v_\kappa^\infty(\theta) \triangleq v_\kappa^\infty(\min\{K_\kappa, \theta+1\}) - v_\kappa^\infty(\theta), \forall \theta \in \Theta, \quad (11)$$

which transforms the incentive constraints (9) as follows:

$$0 \leq \delta(1-2\alpha)\left[\Delta v_\kappa^\infty(\theta-1) + \Delta v_\kappa^\infty(\theta+1)\right] - c, \text{ if } \theta \geq h_\kappa + 1,$$
$$0 \leq \delta(1-2\alpha)\sum_{\tilde{\theta}=0}^{h_\kappa}\Delta v_\kappa^\infty(\tilde{\theta}) - c, \text{ if } \theta = h_\kappa.$$
(12)

The following proposition characterizes the structure of the marginal utilities.

**Proposition 3.** The marginal utilities defined in (11) preserves the following properties:

(1) $\Delta v_\kappa^\infty(\theta) > 0$, $\forall \theta \in \Theta$;

(2) $\Delta v_\kappa^\infty(\theta) > \Delta v_\kappa^\infty(\theta+1)$, $\forall \theta \in \{h_\kappa, \ldots, K_\kappa - 1\}$;

(3) $\sum_{\tilde{\theta}=0}^{h_\kappa-1}\Delta v_\kappa^\infty(\tilde{\theta}) > \Delta v_\kappa^\infty(h_\kappa)$. ∎

The impact of Proposition 3 is briefly explained below. Statement (1) proves that a worker's long-term utility increases with his reputation. In this way, a social norm is capable of providing incentives to workers with positive rewards associated with their compliance (although the incentive might not be sufficiently large depending on the cost $c$). Statement (2) shows that the increase on the long-term utility slows down as a worker's reputation increases. Therefore, a worker with a higher reputation is less rewarded by complying with the social norm and thus, he has less incentive to do so. Statement (3) proves that a worker of reputation $h_\kappa$ has the highest incentive, which is $\delta(1-2\alpha)\sum_{\tilde{\theta}=0}^{h_\kappa}\Delta v_\kappa^\infty(\tilde{\theta}) - c$, among all active workers. Combining Statement (2) and (3), we can conclude that it is the incentive of workers of reputation $K_\kappa$, which is the smallest among all active workers, that determines the sustainability of a protocol.

By applying Proposition 3, we have the following theorem establishing a protocol's sustainability conditions on $(h_\kappa, K_\kappa)$ (i.e. how these parameters should be selected by the protocol designer in order to sustain the resulting social norm as an equilibrium).

**Theorem 2 (Equilibrium Conditions of Social Norms).** Given $p$, $c$, $\delta$, and $\alpha$, a social $\kappa = (\sigma, \tau)$ can be sustained as an equilibrium if and only if

(1) Its social threshold $h_\kappa$ is larger than a constant $\bar{h}(p, c, \delta, \alpha)$;

(2) The highest reputation $K_\kappa$ is smaller than a constant $\bar{K}(p, c, \delta, \alpha)$.

*Proof*: See [16, Appendix A]. ∎

Theorem 2 provides a guideline for selecting the parameters $(h_\kappa, K_\kappa)$ of a sustainable protocol. As the proof shows, increasing the social threshold $h_\kappa$ enlarges the gap between the long-term utilities that can be received by active and isolated workers. This provides a larger threat of future punishment for active workers and hence reduces their incentives of deviation. On the other hand, increasing $K_\kappa$ extends the length of the warning window. Since more deviations are needed to trigger the punishment upon workers, their incentives of compliance decrease as a result.

Theorem 2, however, does not explicitly consider the intrinsic constraints imposed on $(h_\kappa, K_\kappa)$. Specifically, if $\bar{h}(p, c, \delta, \alpha) > \bar{K}(p, c, \delta, \alpha)$ or $\bar{K}(p, c, \delta, \alpha) < 1$, there will be no protocol that can be sustained in workers' self-interests. The existence of sustainable protocol in general is impacted jointly by the intrinsic parameters $c$, $p$, $\delta$, and $\alpha$ (since we set $\lambda = 1$). In the following theorem, we establish sufficient and necessary conditions on these parameters for the existence of sustainable protocols.

**Theorem 3 (Existence of Sustainable Protocols).** Sustainable protocols exist if and only if

(1) Given $\delta$ and $\alpha$, the cost-to-price ratio $r \triangleq \frac{c}{p}$ of the worker is lower than a threshold $\bar{r}(\delta, \alpha)$;

(2) Given the cost-to-price ratio $r$ and $\alpha$, the discount factor $\delta$ is higher than a threshold $\underline{\delta}(r, \alpha)$;

(3) Given $r$ and $\delta$, the error probability $\alpha$ is lower than a threshold $\bar{\alpha}(r, \delta)$ and higher than a threshold $\underline{\alpha}(r, \delta)$.

*Proof*: See [16, Appendix A]. ∎

It should be noted from Theorem 3 that it is neither $p$ nor $c$ but the ratio between them that determines the existence of sustainable protocols. Moreover, sustainable protocols exist if and only if a worker is sufficiently patient (i.e. he puts sufficiently large weight in his future utility rather than saving the immediate cost) and the transaction error probability is sufficiently small (i.e. workers can be correctly rewarded and punished with truthful reports from requesters). However, when $\alpha$ approaches 0, $\Delta v_\kappa^\infty(K_\kappa)$ also approaches 0 and hence workers of reputation of reputation $K_\kappa$ cannot be incentivized to comply with the social norm. Therefore, $\alpha$ is also lower-bounded to ensure the existence of sustainable protocols.

As a corollary of Theorem 3, it can be shown that sustainable protocols always exist in an idealized scenario, with $r \to 0$ and $\delta \to 1$.

**Corollary 1**. There is always at least one sustainable protocol if $r \to 0$, and $\delta \to 1$.

*Proof*: See [16, Appendix A]. ∎

Based on Theorem 2 and 3, an algorithm to design the optimal sustainable protocol can be proposed, which is greatly simplified by applying the threshold-based properties of parameters. The detailed design of the algorithm is deferred to [16].

## IV. PROTOCOL DESIGN FOR REVENUE MAXIMIZATION

In this section, we consider a different protocol design problem when the protocol designer wants to maximize the revenue of the website from all transactions.

The revenue that the website receives from each transaction is $(1-\lambda)p$. It is always to the best interest of the protocol designer to set $\lambda$ as small as possible to unilaterally increase the revenue. However, as shown in Proposition 1, a small $\lambda$ reduces the benefit of a worker, which in turn harms the worker's incentive to comply with the social norm. With more workers being isolated, the number of on-going transactions in each period decreases and the total revenue of the website might also decrease.

Hence, the protocol designer also has to find the balance between the revenue received by the website per transaction and workers' incentives of contribution in the revenue maximization problem. It is worth noting that different from the social welfare maximization where the protocol should maximize workers' contributions in transactions, the revenue maximization only have to minimize the fraction of workers being isolated. This is due to the fact that the website can receive its part of the payment as long as a transaction is initiated (i.e. a worker and a requester is matched), regardless of whether the worker contributes efforts or not afterwards. Hence, the revenue in principle does not necessarily achieve its optimum under a sustainable protocol. However, it will be shown in our simulation that the maximum revenue that can be achieved with sustainable protocols is usually higher than that can be achieved with unsustainable protocols, except in some extreme scenarios. In other words, the optimal revenue of the website is generally achieved with a sustainable protocol in most cases. Therefore, we focus on the revenue maximization among sustainable protocols in this section, which is formalized as follows:

$$\max_{(K_\kappa, h_\kappa, \lambda)} R_\kappa = \sum_{\theta \geq h_\kappa} \eta_\kappa(\theta)(1-\lambda)p$$

subject to

$c \leq \delta(1-2\alpha)\left[v_\kappa^\infty(\min\{L, \theta+1\}) - v_\kappa^\infty(\theta-1)\right]$, if $\theta \geq h_\kappa + 1$,

$c \leq \delta(1-2\alpha)\left[v_\kappa^\infty(\min\{L, \theta+1\}) - v_\kappa^\infty(0)\right]$, if $\theta = h_\kappa$.

(13)

The parameters $(h_\kappa, K_\kappa)$ have a similar impact on the protocol design as in Section III. Particularly, an increasing $h_\kappa$ also reduces $R_\kappa$; whereas an increasing $K_\kappa$ also increases $R_\kappa$. Nevertheless, $\lambda$ plays a different role here. An increasing $\lambda$ increases workers' incentives but reduces the owner's revenue. As a result, the optimal design $\lambda^\#$ for (13) is the minimum value of $\lambda$ that satisfies the incentive constraints (9). In the follow proposition, we characterize how $\lambda^\#$ is influenced by the intrinsic parameters.

**Proposition 4.** (1) $\lambda^\#$ monotonically increases with the cost-to-price ratio $r$ and monotonically decreases with the discount factor $\delta$;

(2) $\lim_{r \to 0} \lambda^\# = 0$, and $\lim_{\delta \to 1} \lambda^\# = r$.

*Proof*: See [16, Appendix B]. ∎

As an immediate result of Proposition 4, we prove in the following proposition that the optimal revenue of (13), denoted as $R^\#$, possesses similar monotonicity results against $r$ and $\delta$.

**Proposition 5.** $R^\#$ monotonically decreases with the cost-to-price ratio $r$ and monotonically increases with the discount factor $\delta$.

*Proof*: See [16, Appendix B]. ∎

Proposition 4 and 5 indicate that with the increase on $r$ and the decrease on $\delta$, the workers' incentives monotonically decrease and hence, $\lambda$ needs be increased by the protocol designer to raise the benefit per transaction of workers at the expense of the revenue of the website.

## V. ILLUSTRATIVE RESULTS

In this section, we provide numerical results to illustrate the features of our proposed protocol design for crowdsourcing websites. A fixed price $p=5$ is utilized, and the cost $c$ depends on the type of tasks but is restricted to be smaller than $p$.

### A. Experiments on social welfare optimization

We first analyze the problem (10) of social welfare optimization. Figure 2 illustrates how the optimal design $(K^*, h^*)$ is influenced by the intrinsic parameters. For illustration purposes, we set $K^* = 0$ and $h^* = 0$ when there is no sustainable protocol.

When the cost and thus the cost-to-price ratio $r$ increases, the workers' incentives decrease. As workers of reputation $K_\kappa$ has the lowest incentive of comply with the social norm, which serves as the bottleneck in the optimal design, $K^*$ monotonically decreases with $r$ as well.

$h^*$ increases at first to provide higher incentives for workers as $r$ increases. However, when $r$ is sufficiently large, $h^*$ starts to decrease due to the intrinsic constraint $h^* \leq K^*$. When $r$ approaches 1, there is no social norm that can be sustained as an equilibrium and hence both $K^*$ and $h^*$ fall to 0. A similar phenomenon is also observed when the discount factor $\delta$ of workers increases from 0 to 1. As workers have more patience, it becomes easier to sustain a protocol where both $K^*$ and the gap $K^* - h^*$ increase, thereby leading to an increased optimal social welfare. It also can be observed that $K^*$ is always higher and $h^*$ is always lower with a smaller transaction error probability $\alpha$, as workers have higher incentives then.

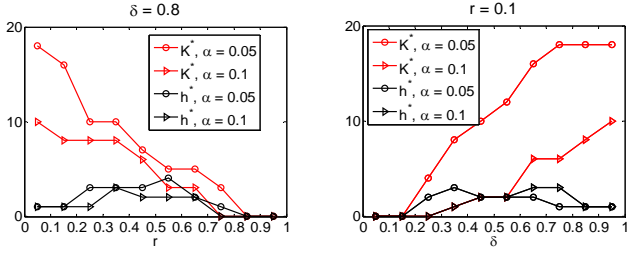

Figure 2  The optimal design $\left(K^*, h^*\right)$

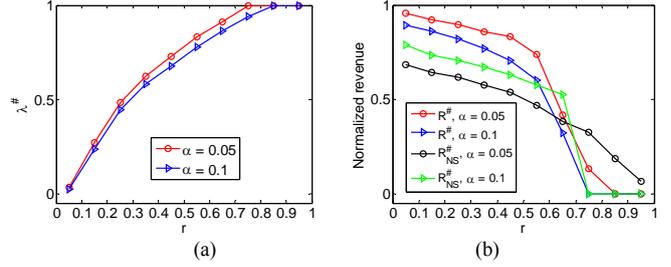

Figure 4  Optimal payment sharing ratio $\lambda^{\#}$ and optimal revenue $R^{\#}$
( $\delta = 0.8$ )

active workers reduces and thus fewer transactions take place in one period and (2) the website owner's revenue share in one transaction decreases. $R^{\#} < R^{\#}_{NS}$ occurs only when $r$ approaches 1. At this region, $R^{\#}$ falls to 0 since no sustainable protocol exists then. Thereby, we show that, with the exception of the scenario in which the cost-to-price ratio is extremely large, $R^{\#}$ is the optimal revenue that can be earned by the website among all protocols.

### C. Experiments with strategic requesters

So far, we consider a fixed pricing scheme and do not explicitly consider the strategic behavior of requesters. In practice, requesters are also rational and self-interested who try to maximize their individual utilities. If the price set for a task is too high, their incentives to submit the payment and participate in the crowdsourcing website will be mitigated.

In this section, we assume that requesters are also strategic and have the freedom in determining whether to pay or not. The utility matrix of a transaction can thus be specified as follows where both workers and requesters are strategic (here we assume $\lambda = 1$):

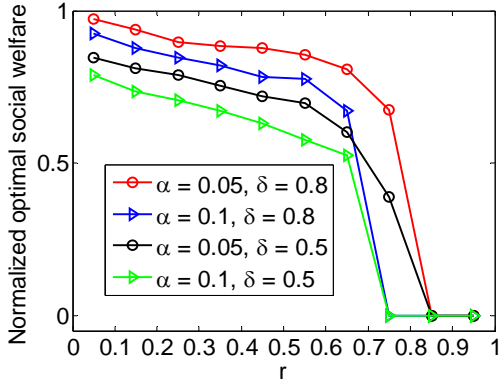

Figure 3  The optimal social welfare

The resulting social welfare according to the optimal design in Figure 2 is plotted in Figure 3, where we set $V = 10$ and normalized the social welfare with its Pareto efficient outcome $V - c$, i.e. the highest social welfare that can be possibly achieved on the website. As the results show, the optimal protocol design leads to a high social welfare which is close to the Pareto efficient outcome when the error probability is small and workers are sufficiently patient.

### B. Experiments on revenue maximization

In this section, we analyze the problem (13) of revenue maximization. It should be noted that (13) only optimally determines the payment sharing ratio under the condition that the resulting protocol is sustainable, i.e. workers can comply with the social norm with their self-interests. In this experiment, we explicitly consider the revenue maximization problem among protocols that cannot be sustained and the related maximum revenue is denoted as $R^{\#}_{NS}$ in order to differentiate with $R^{\#}$.

Figure 4 (a) plots $\lambda^{\#}$ that is the optimal solution of (13). With $r$ increasing, workers have less incentives to comply with the social norm and hence, the website has to share more revenues with workers to incentivize their contributions and $\lambda^{\#}$ monotonically increases. It is also worth noting that $\lambda^{\#}$ is always larger than $r$. Otherwise, the utility that a worker receives from one transaction, which is $\lambda^{\#} p - c$, will be smaller than 0 and they cannot be incentivized regardless of the design of social norms. When $r \to 0$, the cost related to the high effort level of a worker is small and the website can charge more in a transaction with $\lambda^{\#}$ approaching 0. Figure 4 (b) plots the resulting $R^{\#}$ as well as $R^{\#}_{NS}$. Similar to Figure 3, we also normalize the revenues using the highest revenue that can possibly be earned by the website, which is $p$. With $\lambda^{\#}$ increasing against $r$, $R^{\#}$ decreases since (1) the fraction of

Table 2. The utility matrix with strategic requesters

|  |  | Worker | |
|---|---|---|---|
|  |  | H | L |
| Requester | Pay | $V - p$, $p - c$ | $-p$, $p$ |
|  | No pay | $V$, $-c$ | $0$, $0$ |

It is obvious that the dominant action for the requester is "No pay" and hence, an incentive mechanism is also needed to encourage requesters' payments. As discussed in Section II, we also establish a reputation system for requesters as well as the corresponding social norm based incentive protocol. The same forms of the social strategy and the reputation scheme as in (1) and (2) are applied. This is, a requester's reputation increases if he pays a worker the agreed price for a task, and otherwise his reputation will decrease. Once a requester's reputation falls to a threshold value $h_{req}$, one further deviation will trigger an isolation on him for $h_{req}$ periods, during which this requester is forbidden to post tasks. Consequently, a requester has to trade-off his immediate cost, which is $p$, with his future utility, which is proportional to $V$, in order to maximize his long-term utility.

In the experiment, 800 requesters and 200 workers are deployed. We assume that each worker can solve one task per period. Therefore, a requester engages in one transaction every four periods on average. The pricing scheme is still flat-rate, and we examine how the price $p$ influences the social welfare, when the social norms for both workers and requesters are optimally designed. The discount factor $\delta$ is 0.8 for workers and 0.5 for requesters, regarding the fact that requesters usually stay on the website for a shorter period of time. Figure 5 plots the results where the social welfare is normalized with the Pareto efficient outcome $V - c$. With the price $p$ increasing, workers are incentivized to contribute since they receive a higher benefit from each transaction. However, when $p$ is too large, requesters lose their incentives to participate because their cost incurred in each transaction is too high. With the requesters stopping to post tasks, the social welfare decreases. When $p = 10$, which is equal to $V$, few requester will post tasks and the social welfare approaches 0.

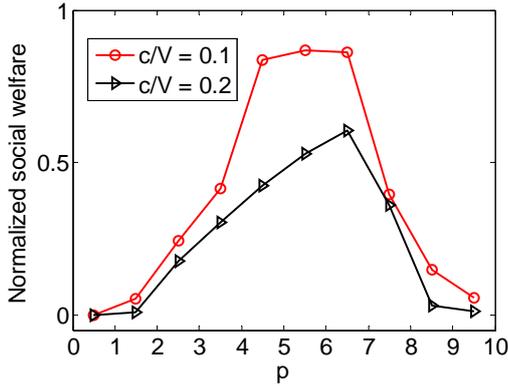

Figure 5  Optimal social welfare with strategic requesters ($\alpha = 0.05$)

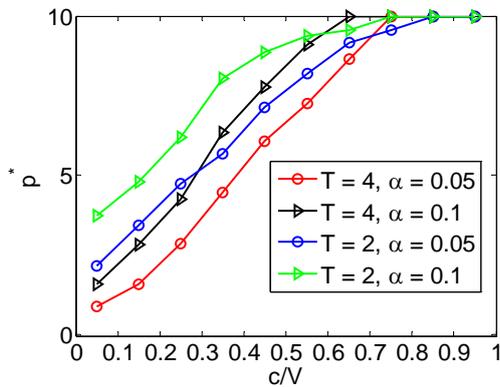

Figure 6  Optimal price $p^*$ ($\delta = 0.8$)

Next, we analyze the design of the optimal price $p^*$ which maximizes the social welfare. Figure 6 illustrates how $p^*$ changes against the ratio $c/V$. With the increase of the cost $c$, more incentives need to be provided to workers and hence, $p^*$ monotonically increases.

We also compare $p^*$ under different values of $\alpha$ and the ratio between the populations of requesters and workers, which is denoted as $T$. A larger $\alpha$ results in lower incentives for workers, which in turn requires a higher price $p^*$ to encourage their contributions. On the contrary, when $T$ becomes larger, requesters have a lower frequency to interact with workers. Therefore, they will put less weight on their future utilities. As a result, $p^*$ should be set smaller in order to reduce requesters' immediate cost to encourage their participations.

## VI. CONCLUSION

In this paper, we build a generic framework to analyze and design incentive protocols based on social norms for crowdsourcing websites. We designed optimal protocols which are sustainable and hence, in which no workers gain by deviating from the social strategy prescribed by the protocol. The structure of optimal protocols maximizing the social welfare is investigated, and we rigorously analyzed the relationships between our designed protocol, the intrinsic parameters (e.g. rewards, costs, workers' patience), and the workers' incentives on a crowdsourcing website. A simple protocol design algorithm is proposed and we proved that our protocol can successfully prevent the "free-riding" problem of workers and incentivize them to contribute their efforts in the task-solving processes. We also explicitly consider the revenue maximization problem of the website owner, and discussed its difference to the problem of the social welfare optimization. Our simulation results illustrate the impacts of the intrinsic parameters and worker' characteristics on the performance of incentive protocols and verify that our social norm based protocol can deliver a good performance which is close to the Pareto efficient outcome. Finally, we considered the incentive problems of requesters and analyze the impact of the pricing scheme on the performance of the crowdsourcing website.